\documentclass[conference,hidelinks]{IEEEtran}
\IEEEoverridecommandlockouts

\usepackage{float}
\usepackage{cite}
\usepackage{amsmath,amssymb,amsfonts}
\usepackage{algorithmic}
\usepackage{graphicx}
\usepackage{textcomp}
\usepackage{xcolor}
\usepackage{fontspec}
\usepackage{polyglossia}

\setmainlanguage{english}
\setotherlanguage{bengali}

\newfontfamily\bengalifont[
    Path=fonts/,
    Script=Bengali,
    Renderer=HarfBuzz
]{NotoSansBengaliUI-Regular.ttf}

\usepackage{doi}
\usepackage{multirow}

\def\BibTeX{{\rm B\kern-.05em{\sc i\kern-.025em b}\kern-.08em
    T\kern-.1667em\lower.7ex\hbox{E}\kern-.125emX}}

\begin{document}

\title{HybridRAG-BN: A Retrieval-Augmented Framework with Fine-Tuned Verification for Bangla KBQA}

\author{
\IEEEauthorblockN{Rathijit Aich, Nirjhar Das, Mahfuzulhoq Chowdhury}
\IEEEauthorblockA{
Department of Computer Science and Engineering\\
Chittagong University of Engineering \& Technology\\
Chattogram, Bangladesh\\
\{aichrathijit, nirjhardasami\}@gmail.com, mahfuz@cuet.ac.bd
}
}

\maketitle

\thispagestyle{plain}

\begin{abstract}
Knowledge-base question answering (KBQA) systems rely on effective retrieval and reasoning mechanisms to generate accurate answers from external knowledge sources. However, developing reliable KBQA systems for low-resource languages such as Bangla remains challenging due to limited retrieval-focused research, scarce language resources, and difficulties in grounding generated responses in external knowledge. In this work, we propose HybridRAG-BN, a retrieval-augmented framework for Bangla KBQA that integrates hybrid retrieval using BM25 and BGE-M3, answer generation using the GGUF version of Gemma-4-31B-Instruct, and a LoRA-fine-tuned Gemma-4-31B-Instruct model for answer verification and refinement. To further improve robustness, the framework incorporates a post-processing stage that addresses unresolved cases through fallback answer replacement and DuckDuckGo-assisted retrieval. Experimental results demonstrate the effectiveness of the proposed framework, achieving token-level F1 scores of 0.71654 and 0.72912 on the public and private leaderboards, respectively, securing first place in the competition.

\end{abstract}

\section{Introduction}

This work was originally developed for the IEEE Computer Society CUET Student Branch Chapter's ML Contest 2.0. Knowledge-base question answering (KBQA) systems aim to answer natural language queries by retrieving and utilizing relevant information from external knowledge sources. Although Retrieval-Augmented Generation (RAG) has significantly improved fact-grounded question-answering in high-resource languages, developing reliable Bangla KBQA systems remains challenging due to limited retrieval-focused research, linguistic complexity, and scarce language resources.

In this work, we present HybridRAG-BN, a retrieval-augmented framework designed to improve the accuracy and robustness of Bangla KBQA. Our system was developed and evaluated on a dataset derived from the Indic-RAG-Suite \cite{ref1}, provided as part of the ``Are You Sure LLM Is Enough? -- Intra CUET ML Contest 2.0 (Advanced Track)'', organized by the IEEE Computer Society CUET Student Branch Chapter. The dataset consists of approximately 3,000 question-answer-reasoning training triples, 1,500 test questions and a knowledge base containing approximately 6,500 Bangla Wikipedia pages.

To address retrieval and reasoning challenges in low-resource settings, the proposed framework consists of four stages:

\begin{itemize}
\item Hybrid retrieval using BM25 and BGE-M3 to combine lexical and semantic matching.

\item Answer generation using the GGUF version of Gemma-4-31B-Instruct.

\item Answer verification and refinement using a LoRA fine-tuned Gemma-4-31B-Instruct model.

\item Post-processing through fallback answer replacement and external DuckDuckGo-assisted retrieval for unresolved cases.

\end{itemize}

\section{Methodology}

\begin{figure}[h!]
    \centering
    \includegraphics[
        width=0.9\columnwidth,
        keepaspectratio
    ]{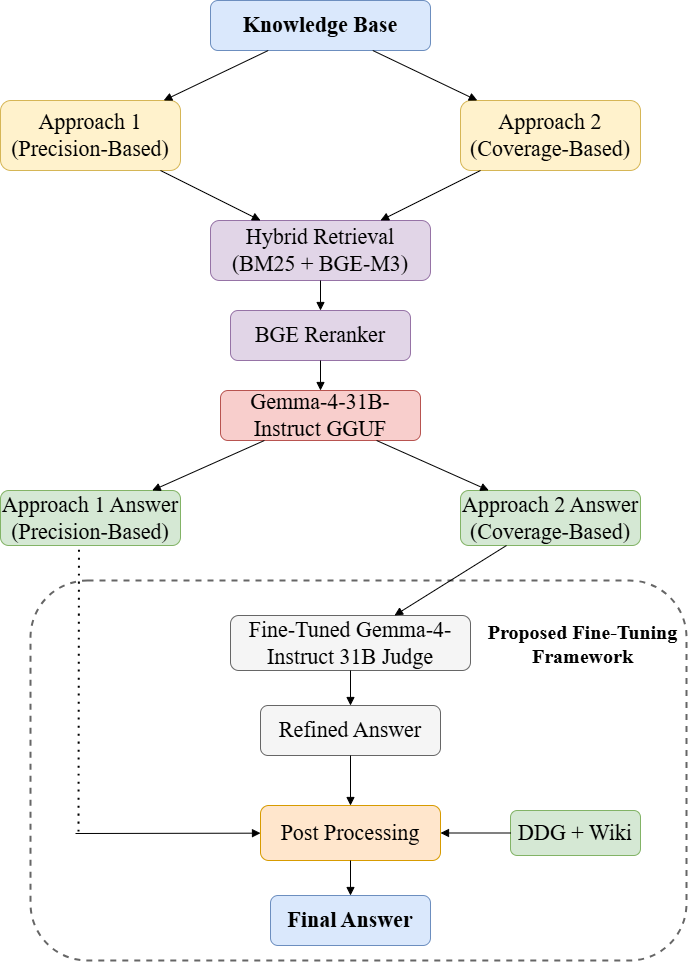}
    \caption{Overview of the proposed question-answering framework.}
    \label{fig:workflow}
\end{figure}

Figure~\ref{fig:workflow} presents an overview of the proposed framework, including the two preprocessing-based approaches, the fine-tuned answer verification module, and the post-processing stage used to generate the final prediction.

\subsection{\textbf{Knowledge Base Preprocessing and Chunking}}

\subsubsection{\textbf{Approach 1 -- Precision-Based}}

The raw Bangla Wikipedia knowledge base was first reconstructed at the page level by splitting the corpus using Wikipedia page markers. Each page was then cleaned by removing recurring boilerplate sections, including page headers, footers, navigation blocks, and language-list sections commonly found in Wikipedia dumps. 

\paragraph{Removed Boilerplate Blocks}

The preprocessing pipeline removed five categories of non-informative content. These included recurring header and footer sections, two variants of tool and navigation blocks commonly embedded within Wikipedia pages, and translation or language-selection blocks. The language-list sections were identified using markers such as ``n {\bengalifont টি ভাষা}'' and {\bengalifont আন্তঃউইকি সংযোগ সম্পাদনা}. Removing these components reduced noise in the knowledge base while preserving the main article content for retrieval.

The cleaned corpus was then divided into overlapping text chunks for retrieval. Chunks were generated with a target length of approximately 1000 characters and an overlap of 150 characters. To preserve semantic coherence, paragraph breaks and predefined boundary markers (such as `` | '' and ``.'') were preferred as chunk boundaries whenever possible. If no suitable boundary was found, a character-based split was applied. The overlap helped preserve contextual information that might otherwise be fragmented across chunk boundaries.

\subsubsection{\textbf{Approach 2 -- Coverage-Based}}

Approach 2 was designed as a less aggressive preprocessing pipeline. In this configuration, only the header block, footer block, and the first type of tool/navigation block were removed.

The cleaned corpus was subsequently divided into overlapping text chunks using the same boundary-aware chunking strategy employed in Approach 1. However, the target chunk size was reduced to approximately 800 characters while maintaining an overlap of 150 characters between consecutive chunks. Table \ref{tab:chunk_stats} shows the comparison.

\begin{table}[h]
\centering
\caption{Chunk Statistics for the Two Preprocessing Approaches}
\begin{tabular}{lccc}
\hline
\textbf{Approach} & \textbf{Chunks} & \textbf{Avg. Length} & \textbf{Max Length} \\
\hline
Approach 1 & 55,182 & 971.6 & 1000 \\
Approach 2 & 75,819 & 772.8 & 800 \\
\hline
\end{tabular}
\label{tab:chunk_stats}
\end{table}

\subsection{\textbf{Retrieval-Augmented Generation (RAG)}}

\subsubsection{\textbf{Dense Embedding Generation}}

Following chunk construction, all knowledge-base chunks were encoded using the BGE-M3 embedding model. The resulting dense embeddings were indexed using FAISS for efficient similarity search. Each chunk was transformed into a normalized dense vector representation, enabling semantic similarity search through vector retrieval.

\subsubsection{\textbf{Hybrid Retrieval}}

To leverage both lexical and semantic matching, a hybrid retrieval strategy was employed. For each query, candidate passages were retrieved using two complementary retrievers. First, BM25 was used to identify chunks with strong keyword overlap. Second, dense retrieval was performed by encoding the query with the BGE-M3 \cite{ref7} embedding model and retrieving the most similar chunk embeddings using vector similarity search.

The candidate passages obtained from both retrievers were combined through weighted score fusion. Retrieval scores were normalized and merged using predefined weights, allowing both lexical relevance and semantic similarity to contribute to the final ranking.

\subsubsection{\textbf{Cross-Encoder Reranking}}

The candidate passages obtained from the hybrid retrieval stage were further refined using the BGE-Reranker-v2-M3 \cite{ref6} cross-encoder model. Unlike the initial retrieval stage, which independently represents queries and passages, the cross-encoder jointly processes each query--passage pair and directly estimates their relevance. Each retrieved candidate was paired with the input query and assigned a relevance score by the reranker. The candidates were then sorted according to these scores, and the highest-ranked passages were selected for context construction.

\subsubsection{\textbf{Context Construction and Augmentation}}

To preserve local contextual information, the passage immediately preceding each selected chunk was included alongside the selected chunks.

\subsection{\textbf{Inference}}

\subsubsection{\textbf{Approach 1 -- Precision-Based}}

As the name suggests, this combines the retrieval pipeline described in the previous section with the Gemma-4-31B-Instruct language model for answer generation. This approach was designed to prioritize answer precision and minimize hallucination by pairing the aggressively cleaned knowledge base with a context-grounded inference strategy. The answers generated by this approach were used in subsequent post-processing.

\paragraph{Language Model}

 The model configuration and inference parameters used throughout the experiments are summarized in Table~\ref{tab:gemma_config}.

\begin{table}[h]
\centering
\caption{Gemma-4-31B-Instruct Inference Configuration}
\label{tab:gemma_config}

\begin{tabular}{lccc}
\hline
\textbf{Parameter} & \textbf{Value} \\
\hline
Model & Gemma-4-31B-Instruct   \\
Checkpoint & \texttt{unsloth/gemma-4-31B-it-GGUF} \cite{ref2} \\
Quantization & \texttt{gemma-4-31B-it-Q3\_K\_S.gguf} \\
Framework & \texttt{llama.cpp} \\
Context window & 10,000 tokens \\
Temperature & 0 \\
Top-p & 1 \\
\hline
\end{tabular}

\end{table}

\paragraph{Prompt Design}

A task-specific prompt was designed to encourage context-grounded and extractive answer generation. The model was instructed to output only the final answer span and avoid unnecessary explanations. To reduce hallucinations, the prompt explicitly required the model to return ``Context {\bengalifont -এ তথ্য নেই}'' when sufficient evidence could not be identified from the retrieved context, rather than generating unsupported answers. This conservative design enabled unresolved cases to be identified more reliably during post-processing. The complete prompt template is provided in Appendix \ref{app:prompt_approach1}. During inference, the maximum generation length was set to 150 tokens.

\paragraph{Retrieval Retry Strategy}

A two-stage retrieval strategy was employed during inference. The retrieval configurations used for the initial and retry stages are summarized in Table~\ref{tab:retry_config}.

\begin{table}[h]
\centering
\caption{Retrieval Configuration}
\label{tab:retry_config}

\begin{tabular}{p{2.2cm}cc}
\hline
\textbf{Parameter} & \textbf{Initial} & \textbf{Retry} \\
\hline
BM25 $k$ & 10 & 10 \\
Dense $k$ & 15 & 15 \\
Final $k$ & 6 & 10 \\
BM25 Weight & 0.65 & 0.65 \\
Dense Weight & 0.35 & 0.35 \\
\hline
\end{tabular}

\end{table}

Retry was triggered when the generated answer indicated insufficient contextual evidence, such as ''Context {\bengalifont -এ তথ্য নেই''}, {\bengalifont ''উল্লেখ নেই''}, or related context-missing responses. In such cases, retrieval was repeated using an expanded candidate pool ($k=10$), and the newly constructed context was provided to the language model for a second inference pass.

\subsubsection{\textbf{Approach 2 -- Coverage-Based}}

This employs the same language model and inference configuration as Approach 1. However, it utilizes the knowledge base constructed using the second preprocessing and chunking strategy described in the preprocessing section. 
 Whereas Approach 1 prioritizes answer precision through aggressive corpus cleaning and conservative answer generation, Approach 2 adopts a more permissive strategy intended to maximize answer coverage.

\paragraph{Prompt Design}

A different prompt was developed for Approach 2 to encourage answer generation even when the retrieved context did not contain sufficient information to directly answer the question. Unlike Approach 1, the prompt discouraged abstention-style responses and instead instructed the model to use its pretrained knowledge to generate the most plausible answer when the retrieved context was insufficient. The prompt also instructed the model to avoid responses such as ''Context{\bengalifont -এ উল্লেখ নেই''}, {\bengalifont ''তথ্য নেই''}, or similar context-missing statements, and to return only the information explicitly requested in the question. The complete prompt template used in Approach 2 is provided in Appendix \ref{app:prompt_approach2}. During inference, the maximum generation length was set to 150 tokens.

\paragraph{Retrieval Retry Strategy}

The retrieval configuration and retry mechanism are the same as shown in Table \ref{tab:retry_config}.

\subsection{\textbf{Proposed Fine-Tuning Framework}}

The final HybridRAG-BN framework uses Approach 2 as its primary pipeline, followed by fine-tuned answer verification, Approach 1-based fallback replacement, and DuckDuckGo-assisted retrieval for the remaining unresolved cases. Although retrieval-augmented generation improves answer quality, incorrect or unsupported answers may still be produced. To address this issue, a LoRA-fine-tuned language model was introduced as an answer-verification component. Rather than generating answers from scratch, the model receives the question, retrieved context, and candidate answer produced by Approach~2, and evaluates whether the answer is correct and supported by the available evidence.

\subsubsection{\textbf{Model Selection}}

The proposed framework employs Gemma-4-31B-Instruct as the base model for fine-tuning. The fine-tuning configuration is summarized in Table~\ref{tab:model_config}.

\begin{table}[h]
\centering
\caption{Fine-Tuning Model Configuration}
\label{tab:model_config}
\begin{tabular}{ll}
\hline
Parameter & Value \\
\hline
Base Model & Gemma-4-31B-Instruct \\
Model Repository & \texttt{unsloth/gemma-4-31B-it} \\
Fine-Tuning Method & LoRA \\
Quantization & 4-bit \\
Maximum Sequence Length & 2000 \\
LoRA Rank ($r$) & 16 \\
LoRA Alpha & 32 \\
LoRA Dropout & 0.0 \\
Bias parameters & Not trained \\
Gradient Checkpointing & Enabled \\
Random Seed & 3407 \\
\hline
\end{tabular}
\end{table}

\subsubsection{\textbf{Training Data Construction}}

The training dataset consisted of three fields: a question, its corresponding ground-truth answer, and an associated reasoning annotation. The dataset was divided into training and validation subsets using an 80/20 split with a fixed random seed.

For each training instance, the question and reasoning annotation were incorporated into an instruction-style prompt, while the ground-truth answer served as the target response.

The following prompt template was used during training:

\begin{quote}
{\bengalifont
প্রশ্ন:
}
[Question]

{\bengalifont
ইঙ্গিত:
}
[Reasoning]
{\bengalifont
নির্দেশনা:

- ইঙ্গিত ব্যবহার করে সঠিক উত্তর নির্ধারণ করো

- যতটা সম্ভব পূর্ণ ও নির্ভুল উত্তর লিখবে

- অসম্পূর্ণ উত্তর লিখবে না

- শুধুমাত্র চূড়ান্ত উত্তর লিখবে

- অতিরিক্ত কিছু লিখবে না

- ব্যাখ্যা লিখবে না
}
\end{quote}

\subsubsection{\textbf{Fine-Tuning Procedure}}

Supervised fine-tuning was performed using TRL's \texttt{SFTTrainer} with LoRA adaptation on a 4-bit quantized Gemma-4-31B-Instruct model. Response-only training was employed so that loss was computed only on assistant-generated tokens. The complete training configuration is summarized in Table~\ref{tab:finetune_hyperparams}.

\begin{table}[h]
\centering
\caption{Fine-Tuning Hyperparameters}
\label{tab:finetune_hyperparams}
\begin{tabular}{ll}
\hline
Parameter & Value \\
\hline
Training Method & Supervised Fine-Tuning (SFT) \\
Framework & TRL SFTTrainer \\
Optimizer & AdamW 8-bit \\
Epochs & 2 \\
Learning Rate & $1.5 \times 10^{-5}$ \\
Warmup Ratio & 0.03 \\
Per-device Batch Size & 1 \\
Gradient Accumulation Steps & 8 \\
Effective Batch Size & 8 \\
Maximum Sequence Length & 1000 \\
Weight Decay & 0.01 \\
Learning Rate Scheduler & Cosine \\
Gradient Checkpointing & Enabled \\
Response-only Training & Yes \\
Random Seed & 3407 \\
Training Examples & 2,400 \\
Optimization Steps & 600 \\
Trainable Parameters & 122.4M (0.39\%) \\
\hline
\end{tabular}
\end{table}

\subsubsection{\textbf{Judge-Based Inference}}

The proposed framework uses the answer generated by Approach 2 as a candidate prediction and applies the fine-tuned model as an answer-verification and refinement component. Rather than generating an answer from scratch, the model receives the original question, the retrieved context, and the candidate answer produced by Approach 2.

A task-specific prompt was developed to guide the verification process. The model was instructed to compare the candidate answer against the available context, remove unnecessary details when appropriate, correct unsupported answers, and preserve the original answer when no confident improvement could be made. To encourage concise and precise refinements, the maximum generation length was limited to 30 tokens. The complete prompt is provided in Appendix \ref{app:prompt_fine_tune}.

\subsubsection{\textbf{Post-Processing}}

\paragraph{Merging with Approach 1 Answers}
Following judge-based inference, an additional post-processing stage was applied to the generated predictions. During manual inspection, a small number of answers were found to contain abstention-style responses such as {\bengalifont ''তথ্য নেই''}, {\bengalifont ''উল্লেখ নেই''}, and related variants indicating that the model had failed to produce a meaningful answer.

To address this issue, a rule-based filtering procedure was employed to identify such responses. For affected instances, the corresponding predictions generated by Approach 1 were used as replacement answers. This strategy was motivated by the observation that Approach 1 occasionally produced valid answers for questions where the fine-tuned verification framework remained overly conservative.

\paragraph{DuckDuckGo (DDG) Search with Entity Extraction}
Following the initial replacement procedure, some predictions still contained the response ''Context {\bengalifont -এ তথ্য নেই''}. To address these remaining failures, an additional retrieval-based post-processing stage was applied.

For each affected question, an entity extraction step was first performed using the Gemma-4-31B-Instruct model. The extracted entity was then used as a query for DuckDuckGo search, and the top Wikipedia result returned by the search engine was selected. The corresponding Wikipedia article content was subsequently retrieved and provided to the language model together with the original question. A task-specific prompt was used for entity extraction and answer generation. The complete prompt is provided in Appendix \ref{app:prompt_pp}.

\section{Experiments}

To evaluate the effect of model scale on retrieval-augmented question answering performance, three other Gemma 4 instruction-tuned models were also investigated: \texttt{google/gemma-4-E2B-it}, \texttt{google/gemma-4-E4B-it}, and \texttt{google/gemma-4-26B-A4B-it}. All experiments were conducted using the same retrieval-augmented generation pipeline, retrieval parameters, and prompt configuration. The Approach 2 preprocessing and chunking strategy was used throughout these experiments to ensure a consistent evaluation setting.

Qualitative inspection of the generated answers revealed a clear improvement in answer quality as model size increased. The results show a consistent improvement in both public and private F1 as model scale increases across the evaluated Gemma 4 variants, demonstrating stronger contextual understanding, and were better able to extract relevant information from the retrieved passages.

\section{Results and Analysis}

The evaluation dataset consisted of 1,500 test questions. Competition rankings were determined using two evaluation splits: a public leaderboard based on approximately 69\% of the test set and a private leaderboard based on the remaining 31\%. 

\subsection{\textbf{Model Scaling Experiments}}

Table~\ref{tab:model_scaling} presents the performance of the Gemma 4 models evaluated using the same retrieval-augmented generation pipeline and the Approach~2 preprocessing strategy.

\begin{table}[h]
\centering
\caption{Performance of Different Gemma 4 Models}
\label{tab:model_scaling}

\begin{tabular}{p{3.0cm}p{1.3cm}p{1.3cm}}
\hline
\textbf{Model} & \textbf{Public F1} & \textbf{Private F1} \\
\hline
Gemma-4-E2B-it \cite{ref5} & 0.59634 & 0.60925 \\
Gemma-4-E4B-it \cite{ref4} & 0.60247 & 0.60727 \\
Gemma-4-26B-A4B-it \cite{ref3} & \textbf{0.66393} & \textbf{0.68064} \\
\hline
\end{tabular}

\end{table}

As shown in Table~\ref{tab:model_scaling}, these results indicate a clear positive relationship between model scale and question-answering performance. Based on this trend, Gemma-4-31B-Instruct was selected as the foundation of the proposed framework.

\subsection{\textbf{Overall Performance Comparison}}

We conduct an incremental ablation study to quantify the contribution of each component. Table~\ref{tab:overall_results} presents the performance of the three evaluated approaches using the Gemma-4-31B-Instruct model. Approach~1 corresponds to the aggressively cleaned knowledge base with retrieval-augmented generation, while Approach~2 uses the alternative preprocessing strategy described previously. The proposed framework extends Approach~2 through fine-tuned answer verification and subsequent post-processing stages.

\begin{table}[h]
\centering
\caption{Overall Token-Level F1 Comparison}
\label{tab:overall_results}

\begin{tabular}{p{2.8cm}p{1.3cm}p{1.3cm}}
\hline
\textbf{Method} & \textbf{Public F1} & \textbf{Private F1} \\
\hline
Approach 1 & 0.69329 & 0.69147 \\
Approach 2 & 0.70223 & 0.69901 \\
Approach 2 + Fine-Tuned Verification & 0.71589 & 0.72495 \\
Proposed Framework & \textbf{0.71654} & \textbf{0.72912} \\
\hline
\end{tabular}

\end{table}

Approach~2 achieved higher performance than Approach~1 on both leaderboard splits, indicating that the less aggressive preprocessing strategy improved retrieval effectiveness. The proposed framework further improved performance substantially, increasing the private leaderboard score from 0.69901 to 0.72912 and the public leaderboard score from 0.70223 to 0.71654.

\subsection{\textbf{Effect of Fine-Tuning}}

A comparison between the outputs of Approach 2 and the fine-tuned verification framework revealed that the fine-tuned model modified 48 of the 1,500 generated answers. These modifications included factual corrections, improved entity resolution, and more precise answer extraction. These targeted modifications produced a substantial improvement in token-level F1, suggesting that a small number of high-impact corrections accounted for a disproportionate share of the performance gain. Representative examples of answers that were altered by the fine-tuned model are presented below.

\textbf{Example 1}

Question:
{\bengalifont নাদিয়া টোনচেভা কোন দুটি উপাধি পেয়েছেন?}

Approach 2:
{\bengalifont ফিদে মাস্টার এবং নারী মাস্টার}

Fine-Tuned:
{\bengalifont ফিদে মাস্টার এবং নারী আন্তর্জাতিক মাস্টার}

\vspace{0.5em}

\textbf{Example 2}

Question:
{\bengalifont শার্লি টেম্পলের চলচ্চিত্রগুলো কোথায় মিলিয়ন মিলিয়ন ডলার আয় করেছে?}

Approach 2:
{\bengalifont মার্কিন এবং কানাডায়}

Fine-Tuned:
{\bengalifont মার্কিন যুক্তরাষ্ট্র এবং কানাডায়}

The fine-tuned verification model was configured with a maximum generation length of 30 tokens to encourage concise answer refinement and maximize token-level F1 performance. While this constraint generally improved answer precision, it occasionally resulted in overly shortened answers for list type questions.

\subsection{\textbf{Effect of Post-Processing}}

Predictions containing abstention-style responses such as {\bengalifont ''তথ্য নেই''}, {\bengalifont ''উল্লেখ নেই''}, or related variants were identified. The corresponding answers from Approach 1 were then examined and used as replacements whenever a more informative prediction was available. This procedure affected 11 predictions. Among them, four were correct answers.

Representative examples are shown below.

\textbf{Example 1}

Question: {\bengalifont গ্রেসন ওয়ালার কোন সালে তার শিক্ষকতার চাকরি ছেড়ে দেন?}

Approach 2 + Fine-Tuned:
{\bengalifont তথ্য নেই}

Approach 1 Replacement:
{\bengalifont ২০২১ সালে}

\vspace{0.5em}

\textbf{Example 2}

Question: {\bengalifont নাজিয়াইং মসজিদটি কোন সালে সংরক্ষিত সাংস্কৃতিক নিদর্শন হিসেবে তালিকাভুক্ত করা হয়?}

Approach 2 + Fine-Tuned:
{\bengalifont প্রদত্ত তথ্যে নাজিইং মসজিদটি কোন সালে সংরক্ষিত সাংস্কৃতিক নিদর্শন হিসেবে তালিকাভুক্ত করা হয়েছে তা উল্লেখ নেই}

Approach 1 Replacement:
{\bengalifont ২০২১ সালে}

\vspace{0.5em}

\textbf{Example 3}

Question:{\bengalifont সুধীর চন্দ্র দাস ১৯৭৭ সালের পশ্চিমবঙ্গ বিধানসভা নির্বাচনে কত শতাংশ ভোট পেয়েছিলেন?}

Approach 2 + Fine-Tuned:
{\bengalifont তথ্য নেই}

Approach 1 Replacement:
{\bengalifont ১০.২৬\%}

\vspace{0.5em}

Following the Approach 1 replacement stage, seven predictions still contained the response ''Context{\bengalifont-এ তথ্য নেই''}. These remaining cases were processed using the DuckDuckGo-assisted retrieval pipeline described previously. By leveraging external retrieval through DuckDuckGo and Wikipedia, additional information could be obtained for questions that were not successfully resolved by either the retrieval-based baseline or the fine-tuned verification framework.

Among the seven unresolved instances, five were successfully converted into answer-bearing predictions. Representative examples are shown below.

\textbf{Example 1}

Question:
{\bengalifont হের্টা মুলারের পিতা কী পেশায় নিয়োজিত ছিলেন?}

Before DDG Retrieval:
Context {\bengalifont -এ তথ্য নেই}

After DDG Retrieval:
{\bengalifont ট্রাক ড্রাইভার}

\vspace{0.5em}

\textbf{Example 2}

Question:
{\bengalifont বোধিনাথ ভেলানস্বামী কোথায় সন্ন্যাস দীক্ষা লাভ করেন?}

Before DDG Retrieval:
Context {\bengalifont-এ তথ্য নেই}

After DDG Retrieval:
{\bengalifont শ্রীলঙ্কার আলাভেডিতে} 

The post-processing stage provided a further improvement over the fine-tuned verification framework, increasing the private token-level F1 score from 0.72495 to 0.72912 and the public token-level F1 score from 0.71589 to 0.71654.

\section{Conclusion}

This paper presented a retrieval-augmented framework for Bangla knowledge-base question answering that combines hybrid retrieval, large language model inference, fine-tuned answer verification, and targeted post-processing. Two knowledge-base preprocessing and chunking strategies were investigated, followed by answer generation using Gemma-4-31B-Instruct and answer refinement through a LoRA-fine-tuned verification model. Experimental results showed that the less aggressive preprocessing strategy improved retrieval effectiveness, while the fine-tuned verification and post-processing stages further improved answer quality and overall token-level F1 performance. The final system achieved public and private leaderboard scores of 0.71654 and 0.72912. Future work may explore stronger retrieval methods, adaptive answer-verification strategies, and improved handling of cases where relevant evidence is absent from the available knowledge sources.

\section{Limitations}

The proposed framework has several limitations. The fine-tuned verification model was constrained to a maximum generation length of 30 tokens, which occasionally resulted in truncated answers for list-type questions. Additionally, the framework relies heavily on Gemma-4-31B-Instruct across multiple pipeline stages, making it computationally expensive. Finally, model scaling experiments were conducted exclusively within the Gemma 4 family, leaving the framework's generalizability to other model families such as Llama, Qwen, or Mistral unexplored.

\clearpage
\appendices

\section{Approach 1 Prompt}
\label{app:prompt_approach1}
\subsection{System Prompt}

The following system prompt was used during inference:

\begin{quote}
You are an ultra-precise question-answering assistant. Think carefully before answering. Output only the final answer. Do not show reasoning.
\end{quote}

\subsection{User Prompt}

The retrieved context and input question were incorporated into the following prompt template:

{\bengalifont প্রশ্নের} semantic meaning {\bengalifont বুঝে} Context {\bengalifont থেকে সবচেয়ে উপযুক্ত ও নির্ভুল উত্তর বের করা। প্রশ্নের শব্দ বা বাক্যাংশ} repeat {\bengalifont করবে না --- শুধুমাত্র} answer part {\bengalifont লিখবে।}

\bigskip

{\bengalifont গুরুত্বপূর্ণ:}

\begin{itemize}

\item {\bengalifont যদি প্রশ্ন} specific {\bengalifont নাম, ব্যক্তি, স্থান, তারিখ, সাল, সংখ্যা, পদবি, পুরস্কার, প্রতিষ্ঠান, ঘটনা বা নির্দিষ্ট তথ্য সম্পর্কে হয়, তাহলে শুধুমাত্র} exact answer part {\bengalifont লিখবে --- অতিরিক্ত কোনো শব্দ, ব্যাখ্যা বা পূর্ণ বাক্য লিখবে না।}

\end{itemize}

{\bengalifont উদাহরণ:}

\medskip

Question:

{\bengalifont বাংলাদেশের রাজধানী কী?}

\medskip

Bad Answer:

{\bengalifont বাংলাদেশের রাজধানী ঢাকা।}

\medskip

Correct Answer:

{\bengalifont ঢাকা}

\bigskip

\begin{itemize}

\item {\bengalifont যদি প্রশ্ন ''কেন'', ''কারণ'', ''কীভাবে'', ''কিসের জন্য'' ধরনের হয়, তাহলে পুরো} semantic context {\bengalifont বুঝে সবচেয়ে উপযুক্ত কারণ বা} short reason phrase {\bengalifont লিখবে --- পূর্ণ বাক্য নয়, অতিরিক্ত ব্যাখ্যাও নয়।}

\end{itemize}

Question:

{\bengalifont রাহুলকে কেন স্কুল থেকে বের করে দেওয়া হয়েছিল?}

\medskip

Wrong Answer:

{\bengalifont রাহুলকে কেন স্কুল থেকে বের করে দেওয়া হয়েছিল কারণ সে নকল করতে গিয়ে ধরা পড়ে।}

\medskip

Correct Answer:

{\bengalifont নকল করতে গিয়ে ধরা পড়ায়}

\bigskip

\begin{itemize}

\item {\bengalifont যদি প্রশ্ন বৈজ্ঞানিক, চিকিৎসাবিজ্ঞান, অ্যানাটমি বা প্রযুক্তিগত বিষয়ের হয়, তাহলে} Context{\bengalifont -কে প্রধান উৎস হিসেবে ব্যবহার করবে এবং প্রয়োজন হলে সাধারণ বৈজ্ঞানিক জ্ঞান ব্যবহার করে সংক্ষিপ্ত ও} function-based {\bengalifont উত্তর দিবে --- সংজ্ঞা বা অপ্রয়োজনীয় ব্যাখ্যা নয়।}

\end{itemize}

Question:

{\bengalifont ফুসফুসের প্রধান কাজ কী?}

\medskip

Wrong Answer:

{\bengalifont ফুসফুস মানুষের শ্বাসযন্ত্রের একটি গুরুত্বপূর্ণ অঙ্গ।}

\medskip

Correct Answer:

{\bengalifont অক্সিজেন গ্রহণ ও কার্বন ডাই-অক্সাইড ত্যাগ করা}

\bigskip

\begin{itemize}

\item {\bengalifont যদি প্রশ্ন কোনো ব্যক্তির স্বভাব, আচরণ বা বৈশিষ্ট্য সম্পর্কে হয় এবং} Context{\bengalifont -এ তা সাধারণীকরণ করা যায়, তাহলে সবচেয়ে} natural short trait/behavior phrase {\bengalifont লিখবে।}

\item {\bengalifont উত্তর লেখার সময় সাধারণ বানান, তারিখ, মাস, সংখ্যা, সাল, ব্যক্তি/স্থানের নাম ও প্রচলিত বাংলা রূপ ঠিক রাখবে (যেমন: ফেব্রুয়ারি, জানুয়ারি, ডিসেম্বর, বঙ্গবন্ধু, চট্টগ্রাম)। তবে} meaning {\bengalifont পরিবর্তন করবে না।}

\end{itemize}

Example:

{\bengalifont অক্টোর } → {\bengalifont অক্টোবর}

\bigskip

\begin{itemize}

\item {\bengalifont প্রশ্নে যতটুকু চাওয়া হয়েছে ততটুকুই লিখবে ---} bracket {\bengalifont এ অতিরিক্ত নাম, উদাহরণ, দেশ,} details {\bengalifont বা} list expand {\bengalifont করবে না, যদি প্রশ্নে} explicitly {\bengalifont তা না চাওয়া হয়।}

\end{itemize}

Question:

{\bengalifont কোন মহাদেশে মরুভূমি আছে?}

\medskip

Wrong Answer:

{\bengalifont এশিয়া (সাহারা, গোবি, আরব মরুভূমি)}

\medskip

Wrong Answer:

{\bengalifont এশিয়া, আফ্রিকা, অস্ট্রেলিয়া --- বিভিন্ন দেশে মরুভূমি রয়েছে}

\medskip

Correct Answer:

{\bengalifont এশিয়া, আফ্রিকা, অস্ট্রেলিয়া}

\bigskip

\begin{itemize}

\item {\bengalifont ''কোথায়'', ''কোন স্থানে'', ''কোন অংশে'', ''কোথায় পাওয়া যায়'', ''কোথায় অবস্থিত'' ধরনের প্রশ্নে যদি} Context{\bengalifont -এ বাস্তব/নির্দিষ্ট স্থান, অংশ, অধ্যায়, পরিশিষ্ট, প্রতিষ্ঠান, শহর, দেশ, ওয়েবসাইট বা} location/entity {\bengalifont স্পষ্টভাবে উল্লেখ থাকে, তাহলে সেটিই} answer {\bengalifont হিসেবে দিবে; ''}Context{\bengalifont -এর অমুক লাইনে/অংশে আছে'', ''উল্লেখ করা হয়েছে'', ''দ্বিতীয় লাইনে আছে'' ধরনের} meta answer {\bengalifont দিবে না।}

\end{itemize}

Question:

{\bengalifont জাতিসংঘের পরিবেশ সংক্রান্ত চুক্তিটি কোথায় উল্লেখ আছে?}

\medskip

Context:

{\bengalifont জাতিসংঘের পরিবেশ সংক্রান্ত একটি বৈশ্বিক চুক্তি, যা ১৯৯২ সালে সাক্ষরিত হয়।}

\medskip

Wrong Answer:

Context{\bengalifont -এ উল্লেখ আছে}

\medskip

Wrong Answer:

{\bengalifont উপরের অংশে আছে}

\medskip

Correct Answer:

{\bengalifont জাতিসংঘের পরিবেশ সংক্রান্ত একটি বৈশ্বিক চুক্তিতে}

\bigskip

\begin{itemize}

\item {\bengalifont যদি প্রশ্নে ''কতবার'', ''কয়বার'', ''কতটি'', ''কয়টি'', ''মোট'', ''সংখ্যা'', ''কতজন'', ''কত বছর'' ইত্যাদি} count {\bengalifont বোঝায়, তাহলে} Context{\bengalifont -এর তথ্য গুনে শুধুমাত্র} final {\bengalifont সংখ্যা/পরিমাণ লিখবে --- তালিকা বা সাল} repeat {\bengalifont করবে না;} Context{\bengalifont -এ আলাদাভাবে} count {\bengalifont লেখা না থাকলেও তথ্য থেকে} count {\bengalifont বের করবে।}

\end{itemize}

\bigskip

{\bengalifont কঠোর নিয়ম:}

\begin{enumerate}
\item {\bengalifont শুধুমাত্র} Context {\bengalifont ব্যবহার করবে।}
\item {\bengalifont প্রশ্নের অর্থ গভীরভাবে বুঝবে।}
\item {\bengalifont পুরো} Context {\bengalifont বুঝে উত্তর দেবে।}
\item {\bengalifont উত্তর} sentence {\bengalifont আকারে লিখবে না।}
\item {\bengalifont শুধুমাত্র} answer part / answer span {\bengalifont লিখবে।}
\item {\bengalifont ''ব্রিটিশ আমলে...'', ''প্রাপ্ত তথ্য অনুযায়ী...'', ''}Context {\bengalifont অনুযায়ী...'' এভাবে শুরু করবে না।}
\item {\bengalifont অপ্রয়োজনীয় শব্দ যোগ করবে না।}
\item {\bengalifont প্রশ্ন} repeat {\bengalifont করবে না।}
\item {\bengalifont প্রশ্ন যদি ''কেন'' হয়,} semantic context {\bengalifont বুঝে সবচেয়ে উপযুক্ত কারণ বের করবে।}
\item {\bengalifont উত্তর সংক্ষিপ্ত,} precise {\bengalifont এবং} natural {\bengalifont হবে।}
\item {\bengalifont এক লাইনে শুধু} final answer {\bengalifont দিবে।}
\item Context{\bengalifont -এর বাইরে কিছু বানাবে না।}
\end{enumerate}

\bigskip

Note:

Context{\bengalifont -এ ''৪ টি ভাষা'', ''} n {\bengalifont টি ভাষা'' এরকম} Wikipedia {\bengalifont থেকে} direct copy {\bengalifont করা} translation option {\bengalifont গুলো রয়ে যেতে পারে --- এগুলো} ignore {\bengalifont করিও, এগুলো মূল তথ্য বা} answer{\bengalifont -এর অংশ না।}

\bigskip

Note:

{\bengalifont যদি} Context{\bengalifont -এ প্রশ্নের উত্তর স্পষ্টভাবে না থাকে বা পর্যাপ্ত তথ্য না পাওয়া যায়, তাহলে শুধুমাত্র ''}Context{\bengalifont -এ তথ্য নেই'' লিখবে --- অতিরিক্ত ব্যাখ্যা, অনুমান বা অন্য কোনো বাক্য লিখবে না।}

\bigskip

Context:

\texttt{\{context\}}

\bigskip

Question:

\texttt{\{question\}}

\bigskip

Answer:

\subsection{Inference Configuration}

\begin{itemize}
    \item Model: Gemma-4-31B-Instruct
    \item Checkpoint: \texttt{unsloth/gemma-4-31B-it-GGUF}
    \item Quantized file: \texttt{gemma-4-31B-it-Q3\_K\_S.gguf}
    \item Framework: \texttt{llama.cpp}
    \item Context window: 10,000 tokens
    \item Temperature: 0
    \item Top-p: 1
    \item Maximum generation length: 150 tokens
\end{itemize}

\clearpage
\section{Approach 2 Prompt}
\label{app:prompt_approach2}

\subsection{System Prompt}

The following system prompt was used during inference:

\begin{quote}
You are an ultra-precise question answering assistant. Think carefully before answering. Output only the final answer. Do not show reasoning.
\end{quote}

\subsection{User Prompt}

The retrieved context and input question were incorporated into the following prompt template:

\begin{flushleft}

{\bengalifont প্রশ্নের} semantic meaning {\bengalifont বুঝে} Context {\bengalifont থেকে সবচেয়ে উপযুক্ত ও নির্ভুল উত্তর বের করা। প্রশ্নের শব্দ বা বাক্যাংশ} repeat {\bengalifont করবে না --- শুধুমাত্র} answer part {\bengalifont লিখবে।}

\bigskip

{\bengalifont গুরুত্বপূর্ণ:}

\begin{itemize}

\item {\bengalifont যদি প্রশ্ন} specific {\bengalifont নাম, ব্যক্তি, স্থান, তারিখ, সাল, সংখ্যা, পদবি, পুরস্কার, প্রতিষ্ঠান, ঘটনা বা নির্দিষ্ট তথ্য সম্পর্কে হয়, তাহলে শুধুমাত্র} exact answer part {\bengalifont লিখবে --- অতিরিক্ত কোনো শব্দ, ব্যাখ্যা বা পূর্ণ বাক্য লিখবে না।}

\end{itemize}

{\bengalifont উদাহরণ:}

\medskip

Question:

{\bengalifont বাংলাদেশের রাজধানী কী?}

\medskip

Bad Answer:

{\bengalifont বাংলাদেশের রাজধানী ঢাকা।}

\medskip

Correct Answer:

{\bengalifont ঢাকা}

\bigskip

\begin{itemize}

\item {\bengalifont যদি প্রশ্ন 'কেন', 'কারণ', 'কীভাবে', 'কিসের জন্য' ধরনের হয়, তাহলে পুরো} semantic context {\bengalifont বুঝে সবচেয়ে উপযুক্ত কারণ বা} short reason phrase {\bengalifont লিখবে --- পূর্ণ বাক্য নয়, অতিরিক্ত ব্যাখ্যাও নয়।}

\end{itemize}

Question:

{\bengalifont রাহুলকে কেন স্কুল থেকে বের করে দেওয়া হয়েছিল?}

\medskip

Wrong Answer:

{\bengalifont রাহুলকে কেন স্কুল থেকে বের করে দেওয়া হয়েছিল কারণ সে নকল করতে গিয়ে ধরা পড়ে।}

\medskip

Correct Answer:

{\bengalifont নকল করতে গিয়ে ধরা পড়ায়}

\bigskip

\begin{itemize}

\item {\bengalifont যদি প্রশ্ন বৈজ্ঞানিক, চিকিৎসাবিজ্ঞান, অ্যানাটমি বা প্রযুক্তিগত বিষয়ের হয়, তাহলে} Context{\bengalifont -কে প্রধান উৎস হিসেবে ব্যবহার করবে এবং প্রয়োজন হলে সাধারণ বৈজ্ঞানিক জ্ঞান ব্যবহার করে সংক্ষিপ্ত ও} function-based {\bengalifont উত্তর দিবে --- সংজ্ঞা বা অপ্রয়োজনীয় ব্যাখ্যা নয়।}

\end{itemize}

Question:

{\bengalifont ফুসফুসের প্রধান কাজ কী?}

\medskip

Wrong Answer:

{\bengalifont ফুসফুস মানুষের শ্বাসযন্ত্রের একটি গুরুত্বপূর্ণ অঙ্গ।}

\medskip

Correct Answer:

{\bengalifont অক্সিজেন গ্রহণ ও কার্বন ডাই-অক্সাইড ত্যাগ করা}

\bigskip

\begin{itemize}

\item {\bengalifont যদি প্রশ্ন কোনো ব্যক্তির স্বভাব, আচরণ বা বৈশিষ্ট্য সম্পর্কে হয় এবং} Context{\bengalifont -এ তা সাধারণীকরণ করা যায়, তাহলে সবচেয়ে} natural short trait/behavior phrase {\bengalifont লিখবে।}

\item {\bengalifont উত্তর লেখার সময় সাধারণ বানান, তারিখ, মাস, সংখ্যা, সাল, ব্যক্তি/স্থানের নাম ও প্রচলিত বাংলা রূপ ঠিক রাখবে (যেমন: ফেব্রুয়ারি, জানুয়ারি, ডিসেম্বর, বঙ্গবন্ধু, চট্টগ্রাম)। তবে} meaning {\bengalifont পরিবর্তন করবে না।}

\end{itemize}
Example:

{\bengalifont অক্টোর} → {\bengalifont অক্টোবর}

\bigskip

\begin{itemize}

\item {\bengalifont প্রশ্নে যতটুকু চাওয়া হয়েছে ততটুকুই লিখবে ---} bracket {\bengalifont এ অতিরিক্ত নাম, উদাহরণ, দেশ,} details {\bengalifont বা} list expand {\bengalifont করবে না, যদি প্রশ্নে} explicitly {\bengalifont তা না চাওয়া হয়।}

\end{itemize}

Question:

{\bengalifont কোন মহাদেশে মরুভূমি আছে?}

\medskip

Wrong Answer:

{\bengalifont এশিয়া (সাহারা, গোবি, আরব মরুভূমি)}

\medskip

Wrong Answer:

{\bengalifont এশিয়া, আফ্রিকা, অস্ট্রেলিয়া --- বিভিন্ন দেশে মরুভূমি রয়েছে}

\medskip

Correct Answer:

{\bengalifont এশিয়া, আফ্রিকা, অস্ট্রেলিয়া}

\bigskip

\begin{itemize}

\item {\bengalifont 'কোথায়', 'কোন স্থানে', 'কোন অংশে', 'কোথায় পাওয়া যায়', 'কোথায় অবস্থিত' ধরনের প্রশ্নে যদি} Context{\bengalifont -এ বাস্তব/নির্দিষ্ট স্থান, অংশ, অধ্যায়, পরিশিষ্ট, প্রতিষ্ঠান, শহর, দেশ, ওয়েবসাইট বা} location/entity {\bengalifont স্পষ্টভাবে উল্লেখ থাকে, তাহলে সেটিই} answer {\bengalifont হিসেবে দিবে;} Context{\bengalifont -এর অমুক লাইনে/অংশে আছে'', 'উল্লেখ করা হয়েছে', 'দ্বিতীয় লাইনে আছে' ধরনের} meta answer {\bengalifont দিবে না।}

\end{itemize}

\bigskip

{\bengalifont কঠোর নিয়ম:}

\begin{enumerate}

\item {\bengalifont শুধুমাত্র} Context {\bengalifont ব্যবহার করবে।}
\item {\bengalifont প্রশ্নের অর্থ গভীরভাবে বুঝবে।}
\item {\bengalifont পুরো} Context {\bengalifont বুঝে উত্তর দেবে।}
\item {\bengalifont উত্তর} sentence {\bengalifont আকারে লিখবে না।}
\item {\bengalifont শুধুমাত্র} answer part / answer span {\bengalifont লিখবে।}
\item {\bengalifont 'ব্রিটিশ আমলে...', 'প্রাপ্ত তথ্য অনুযায়ী...',} `Context {\bengalifont অনুযায়ী...' এভাবে শুরু করবে না।}
\item {\bengalifont অপ্রয়োজনীয় শব্দ যোগ করবে না।}
\item {\bengalifont প্রশ্ন} repeat {\bengalifont করবে না।}
\item {\bengalifont প্রশ্ন যদি 'কেন' হয়,} semantic context {\bengalifont বুঝে সবচেয়ে উপযুক্ত কারণ বের করবে।}
\item {\bengalifont উত্তর সংক্ষিপ্ত,} precise {\bengalifont এবং} natural {\bengalifont হবে।}
\item {\bengalifont এক লাইনে শুধু} final answer {\bengalifont দিবে।}
\item Context{\bengalifont -এর বাইরে কিছু বানাবে না।}

\end{enumerate}

\bigskip

{\bengalifont খুব গুরুত্বপূর্ণ:}

\begin{itemize}

\item {\bengalifont যদি} Context{\bengalifont -এ উত্তর স্পষ্টভাবে না থাকে, তাহলে নিজের} pretrained/general knowledge {\bengalifont ব্যবহার করে সবচেয়ে সম্ভাব্য সংক্ষিপ্ত} answer {\bengalifont দিবে; }'Context{\bengalifont -এ উল্লেখ নেই', `তথ্য নেই', 'উত্তর পাওয়া যায়নি', 'প্রদত্ত তথ্য অনুযায়ী' --- এ ধরনের} wording final answer{\bengalifont -এ কখনো ব্যবহার করবে না।}

\item Final answer {\bengalifont এমনভাবে লিখবে যেন } expected  ground-truth answer{\bengalifont -এর সাথে} token overlap  {\bengalifont সর্বোচ্চ হয় --- অতিরিক্ত} token precision {\bengalifont কমায়, } missing token recall {\bengalifont কমায়।}

\item {\bengalifont সবচেয়ে ছোট কিন্তু সম্পূর্ণ} semantic answer {\bengalifont দিবে ---  \bengalifont উদাহরণ বা}extra explanation, prefix, suffix, bracket, expansion {\bengalifont যোগ করবে না।}

\item {\bengalifont যদি} answer {\bengalifont এক বা কয়েকটি শব্দে দেওয়া সম্ভব হয়, তাহলে শুধু সেই} minimal answer span {\bengalifont লিখবে।}

\item multiple possible wording {\bengalifont থাকলে সবচেয়ে} standard, common {\bengalifont এবং} short wording {\bengalifont ব্যবহার করবে।}

\end{itemize}

\bigskip

Note:

Context{\bengalifont -এ ``৪ টি ভাষা'', ``n টি ভাষা''} এরকম Wikipedia translation noise {\bengalifont থাকতে পারে --- এগুলো} ignore {\bengalifont করিও।}

\bigskip

Context:

\texttt{\{context\}}

\bigskip

Question:

\texttt{\{question\}}

\bigskip

Answer:

\end{flushleft}

\subsection{Inference Configuration}

The same inference configuration used in Approach 1 was employed for Approach 2.

\begin{itemize}
    \item Model: Gemma-4-31B-Instruct
    \item Checkpoint: \texttt{unsloth/gemma-4-31B-it-GGUF}
    \item Quantized file: \texttt{gemma-4-31B-it-Q3\_K\_S.gguf}
    \item Framework: \texttt{llama.cpp}
    \item Context window: 10,000 tokens
    \item Temperature: 0
    \item Top-p: 1
    \item Maximum generation length: 150 tokens
\end{itemize}

\clearpage

\section{Approach 3 Prompt (Fine-Tuned Judge)}
\label{app:prompt_fine_tune}
\subsection{Purpose}

The prompt presented in this appendix was used during the judge-based inference stage of the proposed fine-tuning framework. Unlike Approaches 1 and 2, where the language model directly generated answers from retrieved context, the fine-tuned model operated as a post-processing component.

The objective of the prompt was to verify and refine candidate answers generated by the Approach 2 pipeline. For each instance, the model received the original question, the retrieved context, and the candidate answer produced by Approach 2. The model was instructed to preserve correct answers whenever possible and perform modifications only when sufficient contextual evidence supported a refinement.

\subsection{User Prompt}

The retrieved context and candidate answer were incorporated into the following prompt template:

\begin{flushleft}

{\bengalifont তোমার কাজ হলো} Question, Context {\bengalifont এবং} Candidate Answer {\bengalifont দেখে} final competition answer {\bengalifont বের করা।}

\bigskip

{\bengalifont নিয়ম:}

\begin{itemize}

\item Candidate answer {\bengalifont যদি}  already  {\bengalifont ভালো হয়, একই} answer {\bengalifont রাখবে।}

\item Candidate answer unnecessarily {\bengalifont বড় হলে ছোট করবে।}

\item {\bengalifont নতুন} extra detail  {\bengalifont যোগ করবে না।}

\item Bracket {\bengalifont এর ভিতরে} extra  {\bengalifont তথ্য যোগ করবে না।}

\end{itemize}

{\bengalifont উদাহরণ:}

{\bengalifont ভালো: ১৭৭ ফুট}

{\bengalifont খারাপ: ১৭৭ ফুট (৫৪ মিটার)}

\begin{itemize}

\item {\bengalifont যতটা সম্ভব} minimal answer span {\bengalifont দিবে।}

\item {\bengalifont তবে} answer incomplete {\bengalifont করা যাবে না।}

\item Context {\bengalifont অনুযায়ী} candidate answer verify {\bengalifont করবে।}

\item Context mismatch {\bengalifont হলে} answer {\bengalifont ঠিক করবে।}

\item Question repeat {\bengalifont  করবে না।}

\item Explanation {\bengalifont দিবে না।}

\item Sentence {\bengalifont লিখবে না।}

\item {\bengalifont শুধুমাত্র} final answer {\bengalifont দিবে।}

\end{itemize}

\bigskip

{\bengalifont খুব গুরুত্বপূর্ণ:}

\begin{itemize}

\item {\bengalifont যদি} Context {\bengalifont দিয়ে} answer confidently verify {\bengalifont বা} improve {\bengalifont করা না যায়, তাহলে} Candidate Answer exactly {\bengalifont একই রাখবে।}

\end{itemize}

\bigskip

Question:

\texttt{\{question\}}

\bigskip

Context:

\texttt{\{context\}}

\bigskip

Candidate Answer:

\texttt{\{current\_answer\}}

\bigskip

Final Answer:

\end{flushleft}

\subsection{Inference Configuration}

The fine-tuned model was loaded by combining the Gemma-4-31B-Instruct base model with the LoRA adapters obtained during supervised fine-tuning.

\begin{itemize}
    \item Base Model: Gemma-4-31B-Instruct
    \item Model Repository: \texttt{unsloth/gemma-4-31B-it}
    \item Fine-Tuning Method: LoRA
    \item Quantization: 4-bit
    \item Maximum Sequence Length: 10,000
    \item Decoding Strategy: Greedy Decoding
    \item Temperature: 0.0
    \item Top-p: 1.0
    \item Repetition Penalty: 1.05
    \item Maximum generation length: 30 Tokens
\end{itemize}

\subsection{Inference Inputs}

Each inference instance consisted of the following components:

\begin{itemize}
    \item Question
    \item Retrieved Context (Approach 2)
    \item Candidate Answer (Approach 2 Output)
\end{itemize}

The model generated a refined answer which was subsequently post-processed and validated. When refinement failed, the original candidate answer produced by Approach 2 was retained.

\clearpage

\section{DuckDuckGo-Assisted Retrieval Prompts}
\label{app:prompt_pp}
\subsection{Entity Extraction Prompt}

The following prompt was used to extract the primary entity associated with a question. The extracted entity was subsequently used as a DuckDuckGo search query to identify relevant Wikipedia articles.

\begin{flushleft}

{\bengalifont তুমি একজন } expert entity extractor.

\bigskip

{\bengalifont কাজ:}

{\bengalifont প্রশ্নের শুধুমাত্র} MAIN SUBJECT ENTITY {\bengalifont বের করো।}

\bigskip

{\bengalifont অত্যন্ত গুরুত্বপূর্ণ নিয়ম:}

\begin{enumerate}

\item {\bengalifont সবসময়} PERSON / ORGANIZATION / PLACE / THING extract {\bengalifont  করবে যেটা প্রশ্নটি সম্পর্কে।}

\item {\bengalifont কখনো } extract {\bengalifont করবে না:}
\begin{itemize}
\item {\bengalifont উত্তর}
\item {\bengalifont প্রশ্নে জিজ্ঞাসিত } location
\item {\bengalifont তারিখ}
\item {\bengalifont ঘটনা}
\item {\bengalifont }attribute
\end{itemize}

\item {\bengalifont 'কোথায়' প্রশ্নে} location{\bengalifont নয়, প্রশ্নটি কার সম্পর্কে সেই} entity extract {\bengalifont  করবে।}

\item {\bengalifont 'কখন' প্রশ্নে তারিখ নয়, প্রশ্নটি কী সম্পর্কে সেই} entity extract {\bengalifont করবে।}

\item {\bengalifont সম্পূর্ণ}  official {\bengalifont নাম রাখবে।}

\end{enumerate}

\bigskip

Output format:

{\bengalifont ১ম লাইন = বাংলা} entity

{\bengalifont ২য় লাইন =} English entity

\bigskip

{\bengalifont উদাহরণ:}

\medskip

Question:

{\bengalifont ঢাকা বিশ্ববিদ্যালয় কত সালে প্রতিষ্ঠিত হয়?}

\medskip

Answer:

{\bengalifont ঢাকা বিশ্ববিদ্যালয়}

Dhaka University

\medskip

Question:

{\bengalifont রবীন্দ্রনাথ ঠাকুর কোন পুরস্কার পেয়েছিলেন?}

\medskip

Answer:

{\bengalifont রবীন্দ্রনাথ ঠাকুর}

Rabindranath Tagore

\medskip

Question:

{\bengalifont জাতিসংঘের মহাসচিব কে?}

\medskip

Answer:

{\bengalifont জাতিসংঘ}

United Nations

\bigskip

{\bengalifont কঠোর নিয়ম:}

\begin{enumerate}

\item {\bengalifont সবসময় সম্পূর্ণ} FULL NAME {\bengalifont দিবে।}

\item {\bengalifont কখনো নাম সংক্ষিপ্ত করবে না।}

\end{enumerate}

\medskip

CORRECT:

{\bengalifont কেন উইলিয়ামসন}

{\bengalifont সাকিব আল হাসান}

{\bengalifont রবীন্দ্রনাথ ঠাকুর}

{\bengalifont শেখ মুজিবুর রহমান}

\medskip

WRONG:

{\bengalifont উইলিয়ামসন}

{\bengalifont সাকিব}

{\bengalifont রবীন্দ্রনাথ}

{\bengalifont মুজিব}

\bigskip

Question:

\texttt{\{question\}}

\end{flushleft}

\subsection{Answer Generation Prompt}

After retrieval, the retrieved Wikipedia content and the original question were incorporated into the following prompt template to generate an answer from the external context.

\begin{flushleft}

{\bengalifont প্রশ্নের} semantic meaning {\bengalifont বুঝে} Context {\bengalifont থেকে সবচেয়ে উপযুক্ত উত্তর বের করো।}

\bigskip

Question:

\texttt{\{question\}}

\bigskip

Context:

\texttt{\{context\}}

\bigskip

{\bengalifont গুরুত্বপূর্ণ:}

\begin{itemize}

\item {\bengalifont শুধুমাত্র} answer {\bengalifont লিখবে}

\item sentence {\bengalifont না}

\item reasoning {\bengalifont না}

\item explanation {\bengalifont না}

\end{itemize}

\end{flushleft}

\subsection{Inference Configuration}

The same Gemma-4-31B-Instruct model used in Approach 1 was employed during this stage.

\begin{itemize}
    \item Base Model: Gemma-4-31B-Instruct
    \item Checkpoint: \texttt{unsloth/gemma-4-31B-it-GGUF}
    \item Quantized file: \texttt{gemma-4-31B-it-Q3\_K\_S.gguf}
    \item Framework: \texttt{llama.cpp}
    \item Temperature: 0
    \item Top-p: 1
    \item Maximum generation length: 80 Tokens
\end{itemize}

\subsection{Application Scope}

This retrieval procedure was applied only to predictions that still contained the response ''Context{\bengalifont -এ তথ্য নেই''} after the previous post-processing stage. A total of seven such instances were identified and processed using the DuckDuckGo-assisted retrieval pipeline.

For each affected question, an entity was first extracted using the entity extraction prompt. The extracted entity was then used as a DuckDuckGo query, and the top Wikipedia result returned by the search engine was retrieved. The resulting article content was subsequently provided to the language model together with the original question to generate a replacement answer.

Only answers generated from the DuckDuckGo-assisted retrieval pipeline were used during this stage.

\end{document}